\documentclass{article}

\usepackage{geometry}
\usepackage[numbers]{natbib}

\usepackage{amsmath}
\usepackage{amssymb}
\usepackage{latexsym}
\usepackage{dsfont}
\usepackage{mathtools}
\usepackage{float}
\usepackage{caption}
\usepackage{inputenc}
\usepackage{epsfig}
\usepackage{multicol}
\usepackage{graphicx}
\usepackage{url}
\usepackage{subfigure}
\usepackage{multirow}
\usepackage{algorithm2e}

\DeclareMathOperator*{\argmin}{arg\,min}

\title{Thompson Sampling Guided Stochastic Searching on the Line for Deceptive Environments with Applications to Root-Finding Problems\thanks{A preliminary
version of some of the results of this paper appears in the Proceedings of
AIAI'15.}}

\author{Sondre Glimsdal\thanks{This author can be contacted at: Centre for Artificial Intelligence Research (CAIR), University of Agder, Postboks 422, 4604 Kristiansand, Norway.  E-mail: {\tt
sondre.glimsdal@uia.no}.} ~and~Ole-Christoffer Granmo\thanks{Author's status: {\it Professor}. This author can be contacted at: Centre for Artificial Intelligence Research (CAIR), University of Agder, Postboks 422, 4604 Kristiansand, Norway.  E-mail: {\tt
ole.granmo@uia.no}.}}

\begin{document}

\maketitle

\begin{abstract}
The multi-armed bandit problem forms the foundation for solving a wide range
 of on-line stochastic optimization problems through a simple, yet effective
 mechanism. One simply casts the problem as a gambler that repeatedly pulls
 one out of N slot machine arms, eliciting random rewards. Learning of reward
 probabilities is then combined with reward maximization, by carefully
 balancing reward exploration against reward exploitation. In this paper, we
 address a particularly intriguing variant of the multi-armed bandit problem,
 referred to as the {\it Stochastic Point Location (SPL) Problem}. The gambler
 is here only told whether the optimal arm (point) lies to the ``left'' or to
 the ``right'' of the arm pulled, with the feedback being erroneous with
 probability $1-\pi$. This formulation thus captures optimization in
 continuous action spaces with both {\it informative} and {\it deceptive}
 feedback. To tackle this class of problems, we formulate a compact and scalable
 Bayesian representation of the solution space that simultaneously captures
 both the location of the optimal arm as well as the probability of receiving
 correct feedback. We further introduce the accompanying Thompson
 Sampling guided Stochastic Point Location (TS-SPL) scheme for balancing
 exploration against exploitation. By learning $\pi$, TS-SPL also supports
 {\it deceptive} environments that are lying about the direction of the optimal arm.
 This, in turn, allows us to solve the fundamental Stochastic Root Finding
 (SRF) Problem. Empirical results demonstrate that our scheme deals with both
 deceptive and informative environments, significantly outperforming competing
 algorithms both for SRF and SPL.
\end{abstract}

\section{Introduction}
\label{sec:introduction}

Research on the {\it Stochastic Point Location} (SPL) problem
\cite{oommen1997stochastic}  has delivered increasingly efficient
schemes for locating the optimal point on a line. In all brevity, the optimal
point must be found based on iteratively proposing candidate points, with each
candidate revealing whether the optimal point lies to the candidate's left or
to its right. The provided directions can be erroneous, and the goal is to
locate the optimal point with as few non-optimal candidate proposals as possible.
The SPL problem can also be cast as an agent that moves on a line, attempting
to locate a particular location $\lambda^*$. The agent communicates with a
teacher that notifies the agent whether its current location $\lambda$ is
greater or lower than $\lambda^*$. However, the teacher is of a stochastic
nature and with probability $1-\pi$ feeds the agent erroneous feedback.
 
Despite the simplicity of the SPL problem, SPL schemes have provided novel solutions for a wide range of problems. Intriguing applications include
estimation of non-stationary binomial distributions
\cite{yazidi2012novel}, communication network routing
\cite{oommen2007routing}, and meta-optimization \cite{oommen2009novel}.
Furthermore, recent research that addresses the related {\it Stochastic Root-Finding} (SRF) problem provides promising solutions for parameter
estimation, transportation system optimization, as well as supply chain
optimization \cite{chen2001stochastic,pasupathy2011stochastic}.

{\bf State-of-the-art.}
Adaptive Step Searching (ASS)
\cite{tao2013adaptive} is currently the leading approach to solving SPL problems,
although it is outperformed by Hierarchical Stochastic Searching on the Line
(HSSL) \cite{yazidi2012hierarchical} in highly volatile non-stationary
environments \cite{tao2013adaptive}. 
Optimal Computing Budget Allocation (OCBA) has also been applied to SPL \cite{zhang2015solving} and
provides stable solutions while converging slightly slower than ASS.
Unfortunately, these state-of-the-art
schemes fail when noise increases beyond a certain degree, which happens when
the majority of obtained directions mislead rather than guide. Indeed, by
naively following the directions provided under such circumstances, one is
systematically led away from the optimal point. We refer to this kind of problem
environments as {\it deceptive} environments, as opposed to {\it informative} ones, to be
further clarified below.

To the best of the authors' knowledge, the pioneering CPL-AdS \cite{oommen2003learn} scheme was the
first known approach handling deceptive SPL environments. CPL-AdS relies on
two consecutive phases. In the first phase, a sequence of intelligently
selected questions is used to classify the environment as either informative
or deceptive. By spending a sufficient amount of time in this phase, the
classification can be made arbitrarily accurate. In the second phase, a
regular SPL scheme is applied, except that the directions obtained are
reversed if the problem environment was classified as deceptive in the first phase.
This means that the scheme may have to remain in the first phase for an
extensive amount of time to ensure that the problem environment is correctly
classified, otherwise, one risks being systematically mislead in the second
phase. These properties largely render CPL-AdS inappropriate for on-line or
any-time problem solving.

Recently, HSSL has been extended by Zhang et al. to cover both
informative and deceptive environments, using a Symmetric HSSL (SHSSL)
\cite{zhang2016symmetrical}. This scheme essentially runs two HSSL schemes
in conjunction: one regular that handles informative environments and
one where all feedback from the environment is inverted to handle deceptive
environments. The hierarchy navigation capabilities of HSSL are then exploited
to allow SHSSL to switch between the two HSSLs, depending on the nature of the
environment. However, a significant limitation of HSSL, namely, that $\pi$ must be
larger than the conjugate of the golden ratio, carries over to SHSSL. Indeed,
SHSSL fails to converge for $\pi \in [0.382, 0.682]$, which amounts to
approximately $30\%$ of the feasible values for $\pi$. This is in contrast to
the approach we propose in this paper, as well as to CPL-AdS \cite{oommen2003learn}, since both of these
schemes operate along the whole range of $\pi$ (apart from $\pi=0.5$).

To cast further light on the challenges lined out above, we here
introduce the {\it N-Door Puzzle} as a framework for modeling
deception. We further propose an accompanying novel solution scheme --- {\it
Thompson Sampling guided Stochastic Point Location} (TS-SPL). The TS-SPL
scheme handles both SPL and SRF problems, and is capable of {\it
simultaneously} solving the problem as well as determining whether we are
dealing with an informative or a deceptive environment. As we shall see, not
only does this scheme handle an arbitrary level of noise, but it also
outperforms current state-of-the-art techniques in both informative and deceptive environments.

{\bf The N-Door Puzzle.} In the book "To Mock a Mockingbird" \cite{smullyan1985} the following
puzzle is formulated: \emph{"Someone was sentenced to death, but since the
king loves riddles, he threw this guy into a room with two doors. One leading
to death, one leading to freedom. There are two guards, each one guarding one
door. One of the guards is a perfect liar, the other one will always tell the
truth. The man is allowed to ask one guard a single yes-no question and then
has to decide, which door to take. What single question can he ask to
guarantee his freedom?" } To avoid spoiling the puzzle for the reader, we omit
the solution here and note that asking a double negative question will often
be the correct course of action for these types of puzzles.

The above puzzle can be generalized by increasing the number of doors. Instead of
deciding between merely two doors, the prisoner now faces $N$ doors, with a
guard posted between each pair of doors. Only a single door leads to freedom, the
remaining doors lead to death. At sunrise each day the prisoner is allowed to
ask one of the guards whether the door leading to freedom is to the guard's
left or to the guard's right. However, only a fixed proportion of the guards
answers truthfully, the rest are compulsive liars. Further, the guards are
randomly assigned a position each sunrise, and thus, knowing who lies and who
tells the truth is impossible. As an additional complication, depending on the
mood of the king, the prisoner may be ordered to walk through one door of his
choosing at an arbitrary day. Therefore, to save his life, it is imperative
that the prisoner as quickly as possible determines which door leads to
freedom.

Specifically, let $\pi = \frac{\mbox{\#truthful guards}}{\mbox{\#guards}}$ be
the fraction of truthful guards. Since the guards are randomly assigned a position each
day, the probability of obtaining a truthful answer is governed by $\pi$. If $\pi < 0.5$
then the majority of the guards are compulsive liars, and the guards as an
entity can be characterized as being \emph{deceptive}. Conversely, if $\pi > 0.5$ then the
majority of the guards are truthful and the guards can be seen as
\emph{informative}. For completeness, we mention that the puzzle is unsolvable
for the case where $\pi$ is exactly equal to $\frac{1}{2}$, since it then becomes
impossible to gain information on neither the nature of the doors or the
guards.

{\bf Thompson Sampling.} The Thompson Sampling (TS) principle was introduced
by Thompson already in 1933 \cite{thompson1933} and now forms the basis for several state-of-the-art approaches to the Multi-Armed Bandit (MAB) problem --- a fundamental sequential resource allocation problem that has challenged researchers for decades. At each
time step in MAB, one is offered to pull one out of $N$ bandit arms, which in turn triggers a stochastic reward. Each arm has an underlying probability of providing a reward, however, these probabilities are unknown to the decision maker. The challenge is thus to decide which of the arms to pull at every time step, so as to maximize the expected total number of rewards obtained \cite{bubeck2012regret}.

In all brevity, TS seeks to achieve the above goal by quickly shifting from exploring reward probabilities to maximizing the number of rewards obtained. This is achieved by recursively estimating the underlying reward probability of each arm, using Bayesian filtering of the rewards obtained thus far. TS then simply selects the next arm to pull based on the Bayesian estimates of the reward probabilities (one reward probability density function per arm).

The arm selection strategy of TS is rather straightforward, yet surprisingly efficient. To determine which arm to pull, a single candidate reward probability is sampled from the probability density function of each arm.  \emph{The arm whose sampled value is the highest is the one
pulled next}. The outcome of pulling that arm is in turn used to
perform the next Bayesian update of the arm's reward probability estimate. It is this simple scheme that makes TS
select arms with frequency proportional to the posterior probability of being optimal, leading to quick convergence towards always selecting the optimal arm.

TS has turned out to be among the top performers for traditional MAB problems
\cite{granmo2010tsla,chapelle2011empirical}, supported by theoretical regret
bounds \cite{agrawal2011analysis,agrawal2013further}.
It has also been been successfully applied to contextual MAB problems
\cite{agrawal2013thompson}, Constrained Gaussian Process optimization
\cite{glimsdal2013gaussian}, Distributed Quality of Service Control in Wireless
Networks \cite{granmo2013accelerated}, Cognitive Radio Optimization
\cite{jiao2015optimizing}, as well as a foundation for solving the Maximum a
Posteriori Estimation problem \cite{tolpin2015maximum}.

{\bf Pure Exploration Bandits.}
Throughout this paper we assume that each SPL problem potentially takes part in a larger system consisting of multiple SPL problems, and not necessarily operating in isolation. From existing applications in the literature, such as web crawler balancing \cite{granmo2007learning}, it is clear that the value of an SPL scheme
does hinge upon its ability to cooperate and interact with other decision makers. Such cooperation demands predictable behaviour from the individual decision makers, as well as coordinated balancing of exploring new solution candidates against maintaining good solution candidates. Without such an ability, the system as a whole will not be able to systematically move towards the more promising areas of the search space, gradually focusing in on an optimal configuration.
Therefore, in this paper we omit a direct comparison with schemes that rely on a "fixed sampling \emph{then} decide" approach,
such as Unimodal Bandits \cite{jia2011unimodal}.
For the same reason, we will not investigate purely exploitative bandits
\cite{even2006action, jamieson2014lil, audibert2010best, gabillon2011multi, karnin2013almost}, that is, bandits that
have a pre-defined finite time horizon and whose performance is only measured at the end of that
horizon. Consequentially, such algorithms are free to explore without any negative impact.
These types of algorithms 
are shown to outperform traditional exploitation-exploration bandits such as TS and UCB  for scenarios where exploitation is not required.\footnote{There also exists a wide spectrum of techniques and schemes in 
the literature on the topic of searching with noise. See for instance \cite{pelc2002searching} for a comprehensive survey. These are unable to handle unknown and deceptive environments, with stochastic directional feedback, and are therefore not directly comparable to SPL solution schemes. We have therefore not included this class of techniques in the present paper.}

{\bf Paper Contributions.}
\label{sec:contribution}
The contributions of this paper can be summarized as follows. First of all, we
introduce a novel scheme for solving the SPL problem, namely, {\it Thompson
Sampling guided Stochastic Point Localization} (TS-SPL). First of all, we formulate a
compact and scalable Bayesian representation of the solution space. This Bayesian representation
simultaneously captures both the location of the optimal point (bandit arm) as well as the
probability of receiving correct feedback. We further introduce the
accompanying Thompson Sampling guided Stochastic Point Location (TS-SPL)
scheme for balancing exploration against exploitation. By learning $\pi$, TS-SPL also supports {\it deceptive} environments that are lying about the direction of
the optimal arm. This, in turn, allows us to solve the fundamental Stochastic
Root Finding (SRF) Problem. More specifically, the contributions of the paper can be summarized as follows:
\begin{enumerate}
    \item We introduce the novel TS-SPL scheme that represents the solution space of N-Door Puzzles, and thus SPL problems, in terms of a Bayesian model. As opposed to competing solutions that merely maintain and refine a single candidate solution, our Bayesian model encompasses the complete space of candidate solutions at every time instant. This Bayesian representation of the problem opens up for efficient exploration
        and exploitation of the solution space with Thompson Sampling.
        \item We formulate a compact and scalable Bayesian representation of
the solution space that simultaneously captures both the location of the
optimal point (arm), as well as the probability of receiving correct feedback.
    \item We link TS-SPL to so-called Stochastic Bisection Search; and unify
        accompanying methods under the umbrella of Thompson Sampling.
    \item Similarly, we enhance Soft Generalized Binary Search
        (SGBS), Probabilistic Bisection Search (PBS) and Burnashev-Zigangirov
        Algorithm (BZ) by introducing novel parameter free solutions that take advantage of our Bayesian model of the N-Door Puzzle/SPL problem.
        This approach eliminates previous reliance on prior knowledge of the degree of noise affecting the system to be optimized.
    \item We finally demonstrate the empirical performance of TS-SPL for both
    SPL and SRF problems. TS-SPL outperforms state-of-the-art algorithms in
    both informative and deceptive environments, except that it is beaten by
    the SGBS and BZ schemes with correctly specified observation noise.
\end{enumerate}

% ###################################################################
%
%  #######  ##     ## ######## ##       #### ##    ## ######## 
% ##     ## ##     ##    ##    ##        ##  ###   ## ##       
% ##     ## ##     ##    ##    ##        ##  ####  ## ##       
% ##     ## ##     ##    ##    ##        ##  ## ## ## ######   
% ##     ## ##     ##    ##    ##        ##  ##  #### ##       
% ##     ## ##     ##    ##    ##        ##  ##   ### ##       
%  #######   #######     ##    ######## #### ##    ## ######## 
% 
% ###################################################################

{\bf Paper Outline.}
\label{sec:outline}
The paper is organized as follows. In Section \ref{sec:solution}, we present
our scheme for Thompson Sampling guided Stochastic Point Location (TS-SPL). We
first introduce the Bayesian model of the N-Door Puzzle. Based on the Bayesian
model, we then formulate a TS based scheme that balances solution space exploration against reward maximization. We further extend selected state-of-the-art solution schemes with the Bayesian model that TS-SPL employ. This
extension removes the need for prior information on observation noise. Then, in Section
\ref{sec:results}, we provide extensive empirical results comparing TS-SPL with state-of-the-art schemes for both SPL and SRF. We
conclude in Section
\ref{sec:conclusion} and point to promising venues for further work.

% ###################################################################
%
%  ######   #######  ##       ##     ## ######## ####  #######  ##    ## 
% ##    ## ##     ## ##       ##     ##    ##     ##  ##     ## ###   ## 
% ##       ##     ## ##       ##     ##    ##     ##  ##     ## ####  ## 
%  ######  ##     ## ##       ##     ##    ##     ##  ##     ## ## ## ## 
%       ## ##     ## ##       ##     ##    ##     ##  ##     ## ##  #### 
% ##    ## ##     ## ##       ##     ##    ##     ##  ##     ## ##   ### 
%  ######   #######  ########  #######     ##    ####  #######  ##    ##
%
% ###################################################################
\section{Thompson Sampling guided Stochastic Point Location (TS-SPL)}
\label{sec:solution}

In this section, we introduce the Thompson Sampling guided Stochastic Point Location (TS-SPL) scheme.
At the core of TS-SPL we find a Bayesian model of the N-Door Puzzle (introduced in Section \ref{sec:introduction}).

Formally, we represent an N-Door Puzzle instance as a tuple
$(\lambda^*, \pi^*) \in D \times T$, where $D = \{d_1, \ldots, d_N\}$ is the set of doors and $T \in [0, 1]$ is
the truthfulness of the guards. Let $(\lambda^*, \pi^*)$ be the particular N-Door Puzzle faced. A novel aspect
of TS-SPL is that instead of maintaining a single or a limited set of candidate
solutions, we instead maintain a posterior distribution over the whole solution space, $(\lambda, \pi) \in D \times T$.  This distribution is conditioned
on the feedback already obtained up to time step $n$, allowing us to single in
on $(\lambda^*, \pi^*)$ as the number of time steps increases, ultimately converging to
$(\lambda^*, \pi^*)$.

Assuming no prior information, we assign a uniform distribution over $D \times
T$, i.e., all puzzle instances are equally probable. By gradually refining the posterior distribution over $D \times T$, we can select guards to question in a goal directed manner. In all brevity, we sample a solution candidate $(\lambda^c, \pi^c)$ from $D \times T$,
selecting the guard to the left or to right of $\lambda^c$.
The answer of the selected guard is then used to
update our posterior distribution. By repeating this procedure, the expected
probability of the underlying N-Door Puzzle instance, $(\lambda^*, \pi^*)$, increases monotonically, reducing the probability of
other puzzle instances. In effect, given enough iterations, TS-SPL will
correctly identify the door leading to freedom as the posterior probability of
$(\lambda^*, \pi^*)$ approaches unity.

\subsection{Bayesian Model of the N-Door Puzzle}

The main purpose of the Bayesian model is to facilitate efficient
calculation of a posterior distribution over the possible N-Door Puzzle
instances, $D \times T$. Since the prisoner does not initially know which problem
instance he is facing, and since the observations are stochastic, we cast
$D$ and $T$ as two random variables. We further assume that $D$ and $T$
are independent of each other. Furthermore, the information we obtain from
questioning the guards is represented as a set of 
random variables $Q = \{Q_1, \ldots, Q_n\}$, with each random variable $Q_k$ representing the answer
from question $k$. Finally, we assume that the outcomes of the individual questions, $Q_k \in Q$, are independent when conditioned on $D$ and $P$. For each
question $Q_k$, we can then compute the probability of the answer ("left" or "right")
that we received from the guard, as summarized in Table \ref{tab:door-probabilities}.

\begin{table}[H] % makes little sense if given too early
\centering
\caption{Conditional door probabilities}
\label{tab:door-probabilities}
\begin{tabular}{l|l}
\hline
\parbox{3cm}{ Guard to the left of door to freedom:} &
\parbox{5cm}{$P($left $|$ guard, door, $t) = t$ $P($right $|$ guard, door, $t) = 1 - t$} \\ \hline

\parbox{3cm}{Guard to the right of door to freedom:} & 
\parbox{5cm}{$P($left $|$ guard, door, $t) = 1 - t$   $P($right $|$ guard, door, $t) = t$} \\ \hline

\end{tabular}
\end{table}

As an example, assume that the truthfulness of the guards is $t=0.75$. Let
us further for instance solicit the guard to the left of door $d_4$, with the
guard replying that the door leading to freedom lies to his left. We can then
infer that all doors to the left has a probability of $t=0.75$ of
leading to freedom, and all the doors to the right has the probability $1-t =
0.25$ of leading to freedom.

Applying Bayes Theorem to $P(Q|d,t)$, defined in Table \ref{tab:door-probabilities}, we are able to
derive closed-form expressions for the posterior distributions of both $D$ and $T$. The derivation of $P(d \in D | Q)$ follows (the derivation of $P(t \in T|Q)$ is analogous, and are left out here for the sake of brevity):

\begin{align}
P(d \in D | Q) &= \sum_{t \in T} P(d|Q,t)P(t) \\
        &= \sum_{t \in T}\frac{P(Q | d,t)P(d|t)P(t)}{P(Q|t)}\\
        &\propto \sum_{t \in T}P(Q | d,t)P(d|t)P(t)\\
        &\propto \sum_{t \in T}P(Q | d,t)P(d)P(t)\\
        &\propto \sum_{t \in T}\hat{Q} Q^+ P(d)P(t)
\end{align}
Above, $\hat{Q} = \prod_{k=1}^{n-1} P(Q_k | d,t)$ and $Q^+ = P(Q_n|d,t)$ and (2) follows
directly from Bayes Theorem. We obtain (3) by marginalizing out $Q(Q|t)$. Eq. (4) is a result of the independence of $D$ and $T$, and (5) from
the independence between the questions in $Q$. This leads us to the following two equations for updating our
knowledge surrounding both the door probabilities (Eq. \ref{eq:door-eq}) and the truthfulness of the guards (Eq. \ref{eq:thruth-eq}).

\begin{equation}
P(d \in D | Q) \propto \sum_{t \in T}\hat{Q} Q^+ P(d)P(t)
\label{eq:door-eq}
\end{equation}

\begin{equation}
P(t \in T | Q) \propto \sum_{d \in D}\hat{Q} Q^+ P(d)P(t)
\label{eq:thruth-eq}
\end{equation}

\subsection{Guard Selection}
We have now formally established how we can turn information from the guards into a probability distribution over which door leads to freedom. However, as
mentioned previously, we here face a trade-off between exploring different
doors and zeroing in on the best door found so far. To handle this trade-off we model
the door selection as a so-called Global Information MAB (GI-MAB) \cite{atan2015global}.

To decide on what door
to select at each iteration, we solve the GI-MAB by utilizing the principle of TS. 
Here, the selection process is simply to select a random door proportional to the 
probability that that door is the one leading to freedom. Once the door has been selected, we need
to decide which of the guards to query: the guard to the left or to the right of the door selected. We do this by,
again, selecting one of the guards randomly, proportionally to the sum of the probabilities of the
doors next to each guard. Assume for instance that we have three doors $d_k, 1 \le k \le 3$ with probability of leading to freedom: $P(d_1) = 0.1, P(d_2) = 0.2, P(d_3) = 0.7$. Then, according to the TS principle, these are also the probabilities we use to sample a particular door.
Note that since the answer obtained from each guard queried affects the complete probability distribution over $D$ (the probability associated with every door is updated), we have a GI-MAB as opposed to a traditional MAB.

\subsection{Improving State-of-the-Art Schemes with Bayesian Model}
\label{sec:improving-existing}
A main advantage of TS-SPL compared to similar schemes is the utilization of
the Bayesian model that enables TS-SPL to operate without prior problem parameters. Due
to TS-SPL's close connection to the Probabilistic Bisection Search (PBS)
\cite{horstein1963sequential}, Noisy Generalized Binary Search (NGBS)
\cite{nowak2009noisy} and the BZ algorithm \cite{burnashev1974interval}, we
will here utilize our Bayesian TS-SPL model to also make these other schemes parameter free.

\subsubsection*{Probabilistic Bisection Search}
The goal of Probabilistic Bisection Search (PBS)\footnote{In this context this scheme also covers the Stochastic \emph{Binary} Search}
\cite{waeber2013bisection,nowak2009noisy} is to locate an 
unknown point $X^* \in [0,1]$. To acquire intelligence on the location of $X^*$ one
queries an Oracle of the relation between a point $x$ and $X^*$. The oracle responds by informing
whether $x$ is on the left or the right side of $X^*$. If we assume that the Oracle
always tells the truth, then the well known deterministic Bisection Search that halves the search space with each
query can be employed to efficiently find $X^*$. However, in PBS we assume that the Oracle provides correct answers with probability $p \in (0.5, 1.0]$ and erroneous ones with probability $1-p$.

The origin of PBS can be traced to Horstein \cite{horstein1963sequential}.  In PBS a probability distribution is mapped over the search space 
and is gradually updated using a Bayesian methodology under the assumption that the environment noise $p$ is known a-priori. The search space is then continuously explored using the median of the posterior distribution as the point of interest. It has been shown that PBS has a geometric rate of convergence under the latter assumptions \cite{waeber2013bisection}.

As the noise $p$ is assumed given, one can simply invoke Eq.. \ref{eq:pba-update} to calculate the posterior distribution.
\begin{equation}
    P(d\;|\;Q) \propto P(Q\;|\;d)\;P(d)
    \label{eq:pba-update}
\end{equation}

Here $P(Q\;|\;d)$ is the conditional probability of obtaining answer $Q$. That is, for every location $d$ to the left of $X^*$, the probability that the Oracle directs the decision maker to the right is $p$, $P(Q \;|\;d) = p$. And conversely, $P(Q\;|\;d) = 1-p$ for $d$ to the
right of $X^*$.

To explicitly represent PBS' dependence on knowing $p$ beforehand, we can cast Eq. \ref{eq:pba-update} in terms of Eqs. \ref{eq:door-eq} and \ref{eq:thruth-eq}.
The resulting model then becomes identical to TS-SPL, with the major difference that PBS employ
the median to explore the search space. We denote this new and improved scheme PBS-M.

\subsubsection*{Generalized Binary Search}
The Generalized Binary Search (GBS) problem can be formulated as follows \cite{nowak2011geometry,nowak2009noisy}. Consider a collection of unique binary-valued functions $H$ defined on a domain $X$. Each $h \in H$ is defined
as a mapping from $X$ to $\{-1, 1\}$. Assume that there exists an optimal function $h^*\in H$ that produces
the correct binary labeling for each $x \in X$. For each query $x \in X$ the value of $h^*(x)$ is observed, possibly
corrupted by independent binary noise. 
The objective is then to determine the function $h^*$ using
as few queries as possible. In this paper we restrict $H$ to the class of threshold binary functions with the effect of 
turning the GBS into the informative N-Door Puzzle.

If the feedback is noiseless then the problem boils down to the combinatorial problem
of finding an optimal decision tree in the $H$ space,
a problem that Hyafil and Rivest showed to be NP-complete \cite{hyafil1976constructing,nowak2011geometry}.

The Soft-Decision Generalized Binary Search SDGB-Search \cite{nowak2009noisy,nowak2011geometry} is the \emph{state-of-art} algorithm for finding
$h^*(x) \in H$ when probability of binary noise is less than $1/2$, i.e., for informative environments.

Similarly to TS-SPL, SDBG-Search employs a probabilistic model that for time step $n$ assigns a probability $p_n(h)$ to each $h \in H$. However, for each time-step, it decides which $x \in X$ to query next based on a deterministic heuristic:
\begin{equation}
 \argmin\limits_{x \in X} \sum\limits_{h \in H} | p(h) h(x) |
\label{eq:sdgb-heuristic}
\end{equation}

SDGB uses the following equation to determine and update $p_n(h)$ at each time step:
\begin{equation}
    p_{i+1}(h) \propto p_i(h) \beta^{(1 - z_i(h))/2} (1-\beta)^{(1+z_i(h))/2}
\label{eq:sdgb-update}
\end{equation}
Where $z_i(h) = h(x_i) y_i$ and $y_i \in \{-1, 1\}$ is the response from $h^*(x_i)$.
Simplifying Eq. \ref{eq:sdgb-update} we observe that $z_i(h)$ represents an \textit{AND} operator
that takes on the value 1 if $h(x_i)$ is equal to $h^*(x_i)$ and -1 otherwise. Furthermore, we note that
since $z_h(i) \in \{-1,1\}$, then one of $1 - z_i(h)$ and $1 + z_i(h)$ will have to take the value $2$, while the other takes the value $0$.

By applying the transformation $\pi = 1 - \beta$ we can rewrite Eq. \ref{eq:sdgb-update} as:
\begin{equation}
p_{i+1}(h) \propto = \begin{cases}
p_i(h) \times \pi & \text{if } y_i = h^*(x_i) \\
p_i(h) \times (1-\pi) & \text{else} \\
\end{cases}
\end{equation}.

This update scheme is identical to the one found in PBS and thus suffers from the same limitation (noise probability is
assumed known a priori). In the same manner as we enhanced PBS to utilize a prior over the noise, we can enhance
SDGB (using Eq. \ref{eq:door-eq},\ref{eq:thruth-eq}) to become a parameter free scheme, again employing our Bayesian TS-SPL scheme. In the following, we will denote this improved version of SDGB as SDGB-M.

\subsubsection*{Burnashev-Zigangirov Algorithm}
The Burnashev-Zigangirov (BZ) Algorithm \cite{burnashev1974interval} is one of
the most widely used algorithms for
solving the discrete PBS problem and has in particular been employed in the context of
Active Learning \cite{singh2006active,castro2006upper}.
In BZ, we search for a point $\theta^*$ that is located
on a line. This line is discretized into $m$ bins and we are only allowed to query the borders of the
bins for the direction of $\theta^*$. The BZ algorithm suffers from the same practical limitation as
PBS and SDGB, namely a dependency on knowing the noise level beforehand.

We will now show how BZ can
be improved in a similar fashion as PBS and SDGB, leveraging our Bayesian model. Let $a_i(j)$ denote the probability of $\theta^*$ residing in bin $I_i$ at
time-step $j$. The probability mass function (pmf) of all the bins is
therefore $\textbf{a}(j) = \{a_1(j), a_2(j), \ldots, a_m(j)\}$ with
its cumulative density function (cdf) denoted as $\textbf{A}(j)$.

To decide which point to investigate next, that is, deciding a value for $X_{j+1}$, BZ selects
one of the two closest points to the median of $\textbf{a}(j)$. We denote this point $k = k(j+1)$.
The binary response variable $Y_{j+1} = \mathds{1}\{X_{(j+1)} \ge \theta^*\}$ is
observed with probability $1-\alpha$, whereas
$Y_{j+1} = \mathds{1}\{X_{(j+1)} < \theta^*\}$ with probability $\alpha$.

To update the probability distribution over $\textbf{a}(j)$ we define $\alpha$ as the probability for noise and let 
$\beta = 1 - \alpha$ and $\tau = 2A(k(j+1)) - 1$.

For $i \leq k$ we have
\begin{align*}
    a_i(j+1) = a_i(j) &
        \begin{cases} 
            \frac{2\alpha}{1-\tau(\beta-\alpha)} & \text{if } Y_{j+1} = 0 \\
            \frac{2\beta}{1+\tau(\beta-\alpha)} & \text{if } Y_{j+1} = 1 \\
        \end{cases}
\end{align*}
and for $i > k$
\begin{align*}
    a_i(j+1) = a_i(j) &
        \begin{cases} 
            \frac{2\beta}{1-\tau(\beta-\alpha)} & \text{if } Y_{j+1} = 0 \\
            \frac{2\alpha}{1+\tau(\beta-\alpha)} & \text{if } Y_{j+1} = 1 \\
        \end{cases}
\end{align*}

To change BZ into a parameter free scheme we first notice that for any given
noise $t \in T$: $\beta = t$, $\alpha = 1-t$, $\beta-\alpha = 2t-1$ and
$\tau = A_k(j) - (1 - A_k(j))$. After some simple algebraic manipulations, it turns out that the updating scheme of the BZ algorithm is identical to PBS
expect that:
\begin{enumerate}
\item BZ calculates the normalizing factor as a part of the updating rule instead of using the likelihood value, and then later normalizes as PBS does.
\item BZ samples on the interval edges while PBS samples the midpoints of each interval.
\end{enumerate}

To obtain an enhanced parameter free version of BZ, we simply replace $\alpha$ as a pre-determined constant with a prior distribution that we marginalize out using Eq. \ref{eq:door-eq},\ref{eq:thruth-eq}. We denote the resulting scheme BZ-M.

% ###################################################################   
%
% ########  ########  ######  ##     ## ##       ########  ######  
% ##     ## ##       ##    ## ##     ## ##          ##    ##    ## 
% ##     ## ##       ##       ##     ## ##          ##    ##       
% ########  ######    ######  ##     ## ##          ##     ######  
% ##   ##   ##             ## ##     ## ##          ##          ## 
% ##    ##  ##       ##    ## ##     ## ##          ##    ##    ## 
% ##     ## ########  ######   #######  ########    ##     ###### 
%
% ###################################################################
\section{Empirical Results}
\label{sec:results}
In this section we evaluate the performance of TS-SPL empirically, compared
to competing schemes. We investigate both the effect the various parameter
settings has on behavior, as well as the capability of TS-SPL to handle
different applications, including Stochastic Point Location and Stochastic
Root Finding problems. Unless otherwise noted, the empirical results report
the average of 10 000 independent trials.

For some of the applications we investigate here, we do not find any existing
scheme that handles deceptive environments. Instead, the schemes we have identified
assume that feedback is informative on average. To render comparison
fair, we thus introduce TS-SPL-INF, configured with a prior that the feedback
is informative. This also serves to exemplify the power of our Bayesian
approach, because we can leverage from a prior tailored for the task at
hand. Note that this informed prior is equivalent to the priors used for the
other probability theory based schemes, PBS-M and SDGB-M.

Further note that we apply a fixed set of parameter values across the whole
suite of experiments, set to optimize overall performance. For
SHSSL \cite{zhang2016symmetrical} and HSSL \cite{yazidi2012hierarchical} we used
a tree branching factor of $D = 8$, and for ASS \cite{tao2013adaptive} we set
$N_{\text{max}} = 256$ and $N_{\text{min}} = 1$. For
OCBA \cite{zhang2016symmetrical} we set $n0=15$ and $\theta=1/256$. The priors
used for TS-SPL is uniform over the unit interval and is discretized as $|D|
= 201$, and $|T| = 101$. For the informative schemes TS-SPL-INF, PGA-M,
SGDB-M, BZ-M, we utilize the same prior for the doors as for TS-SPL, $|D| = 201$
however, we use an uniform prior over the interval $(0.5, 1]$ for 
truthfulness, with $|T| = 51$.

We will in the following subsections investigate (1) the effect of different
priors on TS-SPL; (2) TS-SPL's ability to identify the nature of the
underlying stochastic environment; (3) the ability to solve the Stochastic
Point Location Problem; and (4) performance on Stochastic root-finding problems
- a particularly intriguing class of deceptive environments that arises
  naturally as a result of the properties of stochastic root finding.

% ###################################################################
%
% ########  ######           ######  ########  ##       
%    ##    ##    ##         ##    ## ##     ## ##       
%    ##    ##               ##       ##     ## ##       
%    ##     ######  #######  ######  ########  ##       
%    ##          ##               ## ##        ##       
%    ##    ##    ##         ##    ## ##        ##       
%    ##     ######           ######  ##        ######## 
%
% ###################################################################

\subsection{Sensitivity to Discretization and Distribution of Prior}
\label{sec:prior-sensitivity}
Although TS-SPL is a parameter free scheme it depends on defining $D \times T$,
the set of all possible N-Door Puzzles, and then formulating a prior distribution over this space.
Since TS-SPL is a discrete scheme, an important question is how does TS-SPL fare under various
level of discretization, that is, how is the TS-SPL performance affected by the cardinality
of $D \times T$.

We define convergence for TS-SPL to an interval $I$ when 95\% of the
 probability mass is contained in the interval, i.e. $P(I | \text{Observed-
 History}) > 0.95$. The measure of interest is then the number of time-steps
 passed before convergence.

From Table \ref{tab:prior-sens-d} we identify that the cardinality of $D$ in fact, does affect the performance of TS-SPL.
As $|D|$ increases so does the time it takes before TS-SPL converge. However, from Table \ref{tab:prior-sens-d}
it is evident that this relationship between convergence time and $|D|$ is not linear, indeed the
increase in convergence time is insignificant even when doubling from 3200 to 6400 possible doors,
suggesting a logarithmic relation between $|D|$ and convergence time.

\begin{table}
\centering

\begin{tabular}{|c|c|c|c|c|c|c|c|}
\hline
$|D|:$         & 100  & 200  & 400  & 800  & 1600 & 3200 & 6400 \\ \hline
Convergence Steps: & 31.4 & 36.0 & 38.9 & 39.4 & 39.3 & 40.2 & 40.9 \\ \hline
\end{tabular}

\caption{Convergence steps for TS-SPL solving the N-Door Puzzle with
         $\lambda^* = 0.15$, $I=\{0.15 \pm 0.01\}$, $T = \{0.8\}$ and $\pi = 0.8$.}
\label{tab:prior-sens-d}
\end{table}

To see how the cardinality of $|T|$ affects performance we gradually increase the
discretization of the interval $[0,1]$. We fix the cardinality of $|D|$ to 100.
Observing Table \ref{tab:prior-sens-t} it is clear that an increase in the discretization
of $|T|$ does not significantly affect performance.
\begin{table}
\centering

\begin{tabular}{|c|c|c|c|c|c|c|c|}
\hline
$|T|:$             & 50  & 100  & 200  & 400  & 800  & 1600 & 3200 \\ \hline
Convergence Steps: & 51.6 & 50.8 & 48.4 & 52.1 & 51.0 & 52.4 & 52.1 \\ \hline
\end{tabular}
\caption{Convergence steps for TS-SPL solving the N-Door Puzzle with
         $\lambda^* = 0.15$, $I=\{0.15 \pm 0.01\}$, $|D| = 101$ and $\pi = 0.8$.}
\label{tab:prior-sens-t}
\end{table}

Another advantage of our Bayesian scheme is the ability to incorporate prior
information to guide the algorithm. On the other hand, specifying an incorrect
prior can deteriorate performance instead of enhancing it. In Table \ref{tab:prior-overview} we give the results for an informed prior over $T$ and $D$.
With the correct underlying values $\lambda^* = \pi = 0.85$, we specify three
types of priors: Correct $\propto N(\mu = 0.85, \sigma = 0.3)$, Incorrect
$\propto N(\mu = 0.15, \sigma = 0.3)$ and Flat (all solutions equally
probable), denoted C, I and F respectively. From Table \ref{tab:prior-overview} we can see the effect of different priors. In brief, having a
correct prior over the doors contributes more to convergence time than
having a correct prior over the truthfulness of the guards. 
The disadvantage of setting an incorrectly biased prior is also evident, as the flat
prior performs better than any combination involving a biased prior.

\begin{table}
\centering

\begin{tabular}{|l|l|c||l|l|c||l|l|c||}
\hline
Door & Truthfulness & Convergence & Door & Truthfulness & Convergence & Door & Truthfulness & Convergence \\ \hline
F & F & 36.4        & C & F & 30.2        & I & F & 46.4        \\ \hline
F & C & 35.7        & C & C & 30.0        & I & C & 45.2        \\ \hline
F & I & 41.2        & C & I & 40.5        & I & I & 113.1       \\ \hline
\end{tabular}

\caption{Convergence steps for TS-SPL solving the N-Door Puzzle with different priors:
         C - Correct Prior, F - Flat Prior, I - Incorrect prior.
         $\lambda^* = 0.85$, $I=\{0.15 \pm 0.01\}$, $|T| = |D| = 101$ and $\pi = 0.85$.
         }
\label{tab:prior-overview}
\end{table}

\subsection{Tracking the Truthfulness of the Environment}
\begin{figure}
  \centering
  \includegraphics[height=60mm]{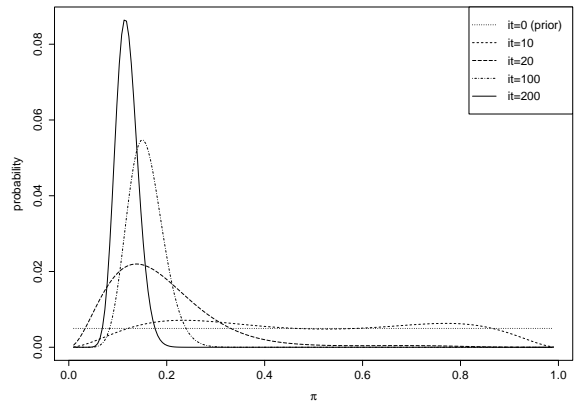}
  \caption{TS-SPL maintains a posterior distribution over $\pi$, here the true underlying value of
  $\pi$ is $0.15$. The figure shows the posterior distribution of $\pi$ after various number of iterations during a 
  single run of TS-SPL. As evident from the figure TS-SPL obtains in this case a sharply peaked posterior over $\pi$
  after only 20 iterations.
  }
  \label{fig:noise-tracker-p20}
\end{figure}

A interesting property of TS-SPL is its ability to provide a distribution over
the truthfulness $\pi$ for that problem instance. This is a significant advantage as
it present the end-user with a better view into the underlying environment when it comes to
practical applications. This can in particular be leveraged in the case
of repeated trials, where the information from previous trials can be us as a
prior on subsequent trials, hence greatly increasing the speed of convergence
as seen in Section \ref{sec:prior-sensitivity}. Figure \ref{fig:noise-tracker-p20} shows the probability of each level of noise as the TS-SPL
progresses with noise probability $\pi = 0.15$  (a highly deceptive environment). As seen, TS-SPL is capable
of quickly estimating the correct value of $\pi$.

% ###################################################################
%
%  ######  ########  ##          ######## ########  ######  ######## 
% ##    ## ##     ## ##             ##    ##       ##    ##    ##    
% ##       ##     ## ##             ##    ##       ##          ##    
%  ######  ########  ##             ##    ######    ######     ##    
%       ## ##        ##             ##    ##             ##    ##    
% ##    ## ##        ##             ##    ##       ##    ##    ##    
%  ######  ##        ########       ##    ########  ######     ##    
%
% ###################################################################
\subsection{Stochastic Point Location}
\label{sec:n-door-result}
The N-Door Puzzle, as outlined in the introduction, is dependent on two variables $\lambda^*$ and $\pi$, 
with $\lambda^*$ specifying the door leading to freedom and $\pi$ the truthfulness of the guards.
Since the N-Door Puzzle does not pose any spatial requirements on the placements of the doors we
can generate a mapping from the N-Door Puzzle to the SPL problem by uniformly placing the 
doors over the unit interval.

As not all of the schemes evaluated in this section are Bayesian, we introduce the notion
of regret, as typical for the multi-armed bandit scenario, as a metric for measuring the performance
of the different schemes.
Regret can be stated as the cumulative penalty from selecting sub-optimal actions. In the case of
SPL we define regret as the (unsigned) distance between the selected point $x$ and the optimal point $\lambda^*$.

\subsubsection{Informative SPL}
We evaluate the performance of TS-SPL and TS-SPL-INF in an informative SPL
problem against algorithms designed to handle informative environments. To the
best of our knowledge this is the first time both the family of PBS based
schemes and the family of SPL based schemes are compared.

The performance of the different schemes is summarized in Table \ref{tab:inf-spl}. One significant observation
is the performance difference between the Learning Automata (LA) based schemes (HSSL and SHSSL) and the Bayesian schemes. 
It is clear that performance wise, Bayesian schemes significantly outperform the LA based schemes, 
however it should be noted that the LA based schemes require less memory and run faster than the
Bayesian ones due to their simplicity.

As can be deduced from Table \ref{tab:inf-spl}, the distance $|\frac{1}{2} - \lambda^*|$ is an important metric for how
hard a particular SPL problem is to solve. This can be explained by the fact that most schemes start exploring
from the center. Thus, if $\lambda^*$ is far from the center then such a scheme
has to obtain more evidence to explore in the peripheral regions of the search
space. This is particularly apparent for PBS-M as its
performance peaks in the case where $\lambda^* = 0.25$, even when faced with
significant noise ($\pi=0.65$).

Since PBS-M pursues the median of the
probability distribution, we can say that PBS-M is conservative in its exploration. This is because it takes significant more evidence to move the point of exploration compared to TS-SPL. TS-SPL on the other hand has a
tendency to explore too much, and as noted by Lattimore
\cite{lattimore2015optimally} using TS for exploration can lead to over-exploration when facing high variance distributions. In the low noise
scenarios, on the other hand, NGBS-M is the most efficient scheme, exploring deterministically.

Moreover, from Table \ref{tab:inf-spl-sd} we observe that TS-SPL-INF exhibits the lowest standard 
deviation overall, and is consequentially the scheme that consistently perform closest to its
expected regret for every trial. This is in sharp contrast to PBS-M who outperform TS-SPL when it comes to average regret, but is unable to do so consistently. NGBS-M also displays significant variance in high noise scenarios.

\begin{table}
\centering
\begin{tabular}{|l|c|c|c|}
\hline
    & Avg Regret $\lambda^*=0.25$ & Avg Regret $\lambda^* = 0.85$ & Avg Regret $\lambda^* = 0.95$  \\ \hline
TS-SPL         & 29.2 / 9.8 / 5.1    & 36.4 / 12.9 / 6.2  & 57.3 / 20.3 / 10.1  \\ \hline
TS-SPL-INF     & 22.2 / 7.3 / 3.7    & \textbf{22.5} / 7.7  / 3.8  & \textbf{23.9} / 8.7 / 4.3    \\ \hline
PBS-M          & \textbf{9.8} / 4.0 / 2.6     & 32.7 / 14.2 / 8.5  & 52.1 / 29.6 / 16.9  \\ \hline
BZ-M           & 23.5 / 5.9 / 2.2    & 27.5 / 6.3 / 2.5   & 35.1 / 9.6 / 3.4    \\ \hline
NGBS-M         & 36.9 / \textbf{3.5} / \textbf{1.0}    & 48.9 / \textbf{4.5} / \textbf{1.5}   & 68.5 / \textbf{7.1} / \textbf{2.3}    \\ \hline
ASS            & 45.8 / 17.0 / 6.7   & 30.4 / 8.9 / 3.6   & 38.8 / 11.7 / 3.9   \\ \hline
OCBA           & 70.8 / 47.4 / 35.2  & 89.9  / 55.8 / 37.1& 112.1 / 78.4 / 48.8 \\ \hline
HSSL           & 117.3 / 23.1 / 8.2  & 111.7 / 16.7 / 4.8 & 131.5 / 19.1 / 5.3  \\ \hline
SHSSL          & 152.2 / 32.6 / 11.8 & 151.8 / 23.5 / 6.5 & 175.1 / 26.1 / 7.3  \\ \hline
\end{tabular}
\caption{Average regret for the different schemes in an informative SPL. The result is reported in the
format "a" / "b" / "c" where $a$ is the average regret for when $\pi=0.65$ and "b" \& "c" is with $\pi=0.75$ and $\pi=0.85$
respectively. The number of time steps per trial is 1000, with 10000 independent trials per data point. }
\label{tab:inf-spl}
\end{table}

\begin{table}
\centering
\begin{tabular}{|l|c|c|c|}
\hline
    & Std. dev. $\lambda^*=0.25$ & Std. dev. $\lambda^* = 0.85$ & Std. dev. $\lambda^* = 0.95$  \\ \hline
TS-SPL         & 16.8 / 5.9 / 2.6   & 20.5 / 6.5 / 3.1   & 30.9 / 10.3 / 4.6  \\ \hline
TS-SPL-INF     & \textbf{13.8} / \textbf{4.2} / 2.0   & \textbf{14.2} / \textbf{4.4} / 2.5   & \textbf{15.7} / \textbf{5.7} / 2.4   \\ \hline
PBS-M          & 15.2 / 10.3 / 10.2 & 69.1 / 40.9 / 31.5 & 94.2 / 71.1 / 56.6 \\ \hline
BZ-M           & 30.4 / 8.9 / 3.1   & 40.8 / 9.8 / 4.9   & 48.8 / 15.3 / 5.3  \\ \hline
NGBS-M         & 68.5 / 8.9 / \textbf{0.9}   & 83.7 / 13.6 / \textbf{1.4}  & 108.7 / 19.4 / \textbf{1.6} \\ \hline
ASS            & 51.6 / 22.4 / 10.1 & 47.8 / 15.7 / 5.4  & 62.3 / 23.1 / 4.6  \\ \hline
OCBA           & 46.2 / 27.6 / 19.4 & 63.9 / 43.9 / 25.6 & 76.1 / 64.6 / 41.9 \\ \hline
HSSL           & 71.7 / 16.1 / 4.6  & 83.6 / 16.1 / 4.2  & 94.8 / 19.4 / 4.5  \\ \hline
SHSSL          & 89.7 / 23.5 / 6.2  & 108.5 / 23.4 / 5.8 & 126.5 / 27.7 / 6.4 \\ \hline
\end{tabular}
\caption{Standard deviation for the different schemes in an informative SPL. The result is reported in the
format "a" / "b" / "c" where $a$ is the standard deviation for when $\pi=0.65$ and "b" \& "c" is with $\pi=0.75$ and $\pi=0.85$
respectively. The number of time steps per trial is 1000 with 10000 independent trials per data point.}
\label{tab:inf-spl-sd}
\end{table}

\subsection{Stochastic Point Location in Deceptive Environments}

With the underlying $\pi$ taking on values in the interval $[0,1]$ we test TS-SPL, CPL-AdS\cite{oommen2003learn} and
SHSSL\cite{zhang2016symmetrical} for speed of convergence and how much regret
on average one accumulates before converging. However, since CPL-AdS operates
in a two-phase mannerm direct comparison with TS-SPL and SHSSL is inappropriate because the latter schemes operate on-line. Oommen et al. states in \cite{oommen2003learn} that this
decision phase needs approximately 200 time steps, and by this time TS-SPL is
already close to converging to the actual solution. To further explore this point, see Table \ref{tab:TS-BN-SPLvsCPL-AdS}. Here it is clear that TS-SPL is superior to CPL-AdS by several orders of magnitude, as well as outperforming SHSSL.

Another interesting observation
is that the performance of TS-SPL is symmetrical around $0.5$. Further note that as stated earlier, SHSSL fails to converge for $\pi
\in [0.382, 0.682]$, so SHSSL is effectively operating with a $30\%$ smaller search space for $\pi$ than both TS-SPL and CPL-AdS.

After modifying PBS, NGBS and BZ to support a Bayesian model of truthfulness, we can use the same prior that we apply in TS-SPL also for these schemes, leading to PBS-M, NGBS-M and BZ-M. The effect of this enhancement to existing schemes is summarized in Table \ref{tab:TS-BN-SPLvsCPL-AdS}.
As clearly seen, the query selection method for these schemes is not suited to handle deceptive environments.

\begin{table}
\centering
\begin{tabular}{|c|c|c|}
\hline
                             & $\pi=0.85$ & $\pi=0.15$ \\ \hline
TS-SPL  ($\lambda^{*}=0.85$) &    6.2         &     \textbf{6.2}      \\ \hline
CPL-AdS ($\lambda^{*}=0.85$) & 501.6 / 354.9  &     842.8/502.3     \\ \hline
PBS-M   ($\lambda^{*}=0.85$) &  31.5          &     77.5     \\ \hline
BZ-M    ($\lambda^{*}=0.85$) &  4.9           &     352.5  \\ \hline
NGBS-M  ($\lambda^{*}=0.85$) &  \textbf{1.4}           &     191.2  \\ \hline
SHSSL   ($\lambda^{*}=0.85$) &  6.5           &     6.5    \\ \hline
\end{tabular}
\caption{Cumulative regret for the deceptive SPL problem.
All entries were estimated taking the average of an ensemble of 10000 independent trials and
each entry corresponds to the estimated cumulative regret after $N=1000$ time steps,
leading to a negligible variance of the estimates relative to
the large difference in performance among the competing schemes.
The number after the slash corresponds to the regret accumulated after CPL-AdS has determined if the teacher
is informative or deceptive.}
\label{tab:TS-BN-SPLvsCPL-AdS}
\end{table}

% ###################################################################
%
% ########   #######   #######  ########    ######## #### ##    ## ########  #### ##    ##  ######   
% ##     ## ##     ## ##     ##    ##       ##        ##  ###   ## ##     ##  ##  ###   ## ##    ##  
% ##     ## ##     ## ##     ##    ##       ##        ##  ####  ## ##     ##  ##  ####  ## ##        
% ########  ##     ## ##     ##    ##       ######    ##  ## ## ## ##     ##  ##  ## ## ## ##   #### 
% ##   ##   ##     ## ##     ##    ##       ##        ##  ##  #### ##     ##  ##  ##  #### ##    ##  
% ##    ##  ##     ## ##     ##    ##       ##        ##  ##   ### ##     ##  ##  ##   ### ##    ##  
% ##     ##  #######   #######     ##       ##       #### ##    ## ########  #### ##    ##  ######   
%
% ###################################################################

\subsection{Stochastic Root-Finding Problem}
The deterministic root finding problem is the procedure of locating a root $x^*$ such
that $g(x^*) = 0$ for a function $g(x)$ defined over an interval $(a,b)$.
We assume that $g(x)$ is unknown, however, an oracle returns $g(x)$ when queried at
point $x$. Then the problem becomes, how can we using as few queries as possible determine
the root $x^*$? If the response from
the oracle is noisy, then we obtain the Stochastic Root Finding Problem (SRFP)\cite{pasupathy2011stochastic}.

One approach to solving the deterministic root finding problem is the Bisection Method.
In the Bisection Method we halves the search space each iteration by continually querying the
oracle on the mid point of the remaining search space. However, for the SRFP the Bisection Method
is unable to discard half of the search space since the oracle may provide false information
regarding value of $g(x)$.

The objective is therefore to select a sequence of queries $x_1, x_2, \ldots$ to 
gather information about $x^*$ such that the final query $x_n$ is close to $x^*$ i.e., 
$|x_n - x^*| < \epsilon$. \cite{waeber2011bayesian}

Formally, let $g: (0,1) \rightarrow \mathbb{R}$ be a function such that given $x^* \in (0, 1)$ then
$\mbox{sign}(g(x^-))$ is equal for all $0 < x^- < x^*$, and $\mbox{sign}(g(x^+)) \neq \mbox{sign}(g(x^-))$ for all
$x^+ > x^* > 1$.

For any $x \in (0,1)$ the oracle generate a sample $Y(x) = g(x) + \epsilon(x)$ where
$\epsilon(x)$ is stochastic noise. Furthermore, define $S(x)$ as the sign of $Y(x)$ i.e.  $S(x) = \text{sign}(Y(x))$.
As we shall see, the TS-SPL as well as PBS and its variants operate solely using $S(x)$. While disregarding the
scalar information of $Y(x)$ might seem wasteful, it open up for the highly efficient Bayesian framework deployed by TS-SPL.
In the scenario that we will investigate here, noise reverse the sign of the function value, i.e., $Y(x)$ returns $g(x)$ with
probability $\pi$ and $-g(x)$ with probability $1-\pi$.

The traditional way of solving SRFP is to apply a variant of
Stochastic Approximation (SA) \cite{robbins1951stochastic,kiefer1952stochastic}.
Implementation wise SA methods\footnote{
    The form of SA shown here is also referred to as Classical Stochastic Approximation (CSA) as it
    closely resemble the original form proposed by Robbins and Monro \cite{pasupathy2010root}.
} extend or modify the iterative Newton-Raphson algorithm to handle noise:
\[
x_{n+1} = x_n - a_n Y_n(x_n)
\]
where $\{a_n\}$ is a sequence of step lengths that decreasing as $n$ increase. Applying SA to SRFP has been 
extensively studied in the literature and it is outside the scope of this article to give a full literature
review, interested readers are referred to \cite{lai2003stochastic,asmussen2007stochastic,pasupathy2011stochastic} and the
references therein. As there exists a myriad of different SA algorithms we have selected one of the more fundamental approaches to form
a basis for comparing the different types of schemes.
Note that a limitation of this SA scheme is that $g(x)$ is required to be monotone.

The main difference between SRFP and SPL is that unlike SPL, SRFP does not provide feedback concerning the
direction of the root $x^*$ from the query location $x$. To map the feedback $S(x) \in \{-1, 1\}$ into a direction it is necessary to know whether $g(x)$ is increasing or decreasing. If it is increasing and $S(x) = 1$ then
the root $x^*$ is to the left of $x$ (and to the right if $S(x)=-1$). Conversely, if $g(x)$ is decreasing and
$S(x) = 1$ then $x^*$ is to the right of $x$ (and to the left if $S(x) = -1$).

Learning the direction of $g(x)$ can be done by repeatedly querying a single point on the edge of the interval $(0,1)$.
To gain an insight into how many repeated samples are sufficient we  employ the two sided Hoeffding's inequality 
$ P(|\bar{X} - E[\bar{X}]| \geq \delta) \leq 2e^{-2n\delta^2} $
where $\bar{X}$ is the average of $n$ queries at $x$, $\delta$ is a
value such that $|\pi - \frac{1}{2}| \geq \delta$. Setting the rhs. equal to $p$ and solving
for $n$, we obtain $n \geq -\frac{\log(p/2)}{2\delta^2}$. Plugging in for $\delta = 0.05$ and $p=0.99$
we obtain $\left \lceil{n}\right \rceil = 62$. Thus we are 99\% sure of our estimate of $g(x)$,
given that $|\pi - 0.5| \geq 0.05$.

However, it turns out that TS-SPL, being able to handle a deceptive environment does not require
this sampling phase. It merely require an arbitrary, yet consistent mapping of each sign to a direction.
For instance, $S(x) = 1 \Rightarrow$ \emph{left}, and $S(x) = -1 \Rightarrow$ \emph{right}.
The reason for this is that if the 
initial mapping is wrong, then TS-SPL will recognize that the feedback is deceptive and
thus still be able to solve the problem with no additional effort. An informative scheme will on
the other hand be unable to recognizing this and will therefore be unable to find the root without
an additional sampling phase.

\begin{figure}
  \centering
  \includegraphics[width=80mm]{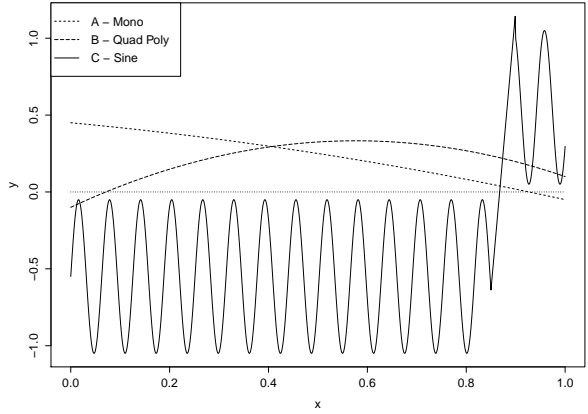}
  \caption{The three functions A, B and C for benchmarking stochastic root finding schemes.}
  \label{fig:threshold-functions}
\end{figure}

We remark that the above sampling procedure \emph{only} enables the other methods to handle SRFP for an unknown
functions in an informative environment, it does not provide a definite answer to whether the environment is informative 
or deceptive. See \cite{oommen2003learn} for a way to implement this as an additional sampling phase.
The functions that we use to measure performance and compare schemes are illustrated in Figure \ref{fig:threshold-functions}.

\begin{table}
\centering
\begin{tabular}{|l|c|c|c|}
\hline 
Func A     & Avg Regret. $\pi = 0.65$ & Avg Regret. $\pi = 0.75$ & Avg Regret $\pi = 0.85$ \\ \hline
TS-SPL     &  46.0  & 17.1  &  8.6  \\ \hline
TS-SPL-INF &  55.2 (24.3)  & 39.1 ( 8.1 )  & 35.0 ( 4.1 )  \\ \hline
PBS-M      &  55.1 (24.2) & 40.3 ( 9.3 )  & 35.2 ( 4.2 )  \\ \hline
NGBS-M     &  53.4 (22.4)  & 35.3 ( 4.3 ) & 32.9 ( 1.9 )   \\ \hline
BZ-M       &  63.8 (32.8)  & 39.9 ( 8.9 ) & 33.8 ( 2.8 )   \\ \hline
SA         &  \textbf{32.4}  & \textbf{14.8} &  \textbf{5.4}   \\ \hline
ASS        &  80.4 (49.4)  & 38.7 ( 7.7 )& 33.5 ( 2.5 )  \\ \hline
HSSL       &  60.4 (29.4) & 45.5 ( 14.6 )& 35.8 ( 4.8 )   \\ \hline
SHSSL      &  62.8  & 20.2 & 6.7 \\ \hline
CPL-AdS    &  162.1 (107.9) &  146.3 (97.4) & 135.3 (90.1) \\ \hline
\end{tabular}
\caption{Average residuals for the different schemes when finding the root of the monotonic function A
under various noise levels. The root is $x^*=0.07104$.
The results are given in the format "average residuals (average residuals after sampling)" for each
scheme. For CPL-AdS the sampling period is the estimation period (epoch 0) as defined by the scheme.
The number of iterations per trial is 250 with 10000 independent trials per data point.}
\label{tab:func-a-results}
\end{table}

\begin{table}
\centering
\begin{tabular}{|l|c|c|c|}
\hline 
Func B     & Avg Regret. $\pi = 0.65$ & Avg Regret. $\pi = 0.75$ & Avg Regret $\pi = 0.85$ \\ \hline
TS-SPL     &  47.1  &  \textbf{17.8}  & \textbf{8.5}   \\ \hline
TS-SPL-INF & 53.8 ( 22.9 ) & 39.5 ( 8.5 ) & 35.1 ( 4.1 )  \\ \hline
PBS-M      & \textbf{41.1} ( 10.2 ) & 35.3 ( 4.31 )& 33.3 ( 2.3 )  \\ \hline
SGBS-M     & 50.6 ( 19.7 ) &35.7 ( 4.7 ) &  33.0 ( 2.0 )  \\ \hline
BZ-M       & 60.3 ( 29.4 ) & 39.6 ( 8.7 ) & 33.6 ( 2.6 )   \\ \hline
SA         & 175.1  & 204.5  &  223.3  \\ \hline
ASS        & 81.6 ( 50.6 ) & 39.5 ( 8.7 ) & 40.2 ( 9.0 ) \\ \hline
HSSL       & 85.3 ( 54.4 ) & 50.6 ( 19.6 ) & 39.0 ( 8.0 )  \\ \hline
SHSSL      &  75.4  &  30.8 & 12.7 \\ \hline
CPL-AdS    &  117.9 (109.3)      &  116.7 (107.1)   &   144.9 (96.5)     \\ \hline
\end{tabular}
\caption{Average residuals for the different schemes when finding the root of the quadric function B
under various noise levels. The root is $x^*=0.9270$.
The results are given in the format "average residuals (average residuals after sampling)" for each
scheme. For CPL-AdS the sampling period is the estimation period (epoch 0) as defined by the scheme.
The number of iterations per trial is 250 with 10000 independent trials per data point.}
\label{tab:func-b-results}
\end{table}

\begin{table}
\centering
\begin{tabular}{|l|c|c|c|}
\hline 
Func C     & Avg Regret. $\pi = 0.65$ & Avg Regret. $\pi = 0.75$ & Avg Regret $\pi = 0.85$ \\ \hline
TS-SPL     & \textbf{36.9}   & \textbf{13.7} & \textbf{6.4}  \\ \hline
TS-SPL-INF & 52.8 ( 21.9 )  & 38.8 ( 7.8 ) &  34.9 ( 3.9 )   \\ \hline
PBS-M      & 49.6 ( 18.7 )  & 39.2 ( 8.3 ) &  34.3 ( 3.3 )   \\ \hline
SGBS-M     & 47.2 ( 16.2 )  & 34.6 ( 3.6 ) &  32.5 ( 1.6 )  \\ \hline
BZ-M       & 58.8 ( 27.9 )   & 38.4 ( 7.4 )  &  33.7 ( 2.7 )     \\ \hline
SA         & 149.0 & 178.0 & 185.0    \\ \hline
ASS        & 54.3 ( 23.4 )  & 39 ( 8.0 )&  33.5 ( 2.5 )   \\ \hline
HSSL       & 75.2 ( 44.3 )  & 44.3 ( 13.4 ) &  35.6 ( 4.6 )  \\ \hline
SHSSL      & 56.5 & 18.4 & \textbf{6.4} \\ \hline
CPL-AdS    & 153.0 (101.6)   & 156.2 (103.7)  & 165.0  (109.6)  \\ \hline
\end{tabular}
\caption{Average residuals for the different schemes when finding the root of the sinusoidal function B
under various noise levels. The root is $x^*=0.8675$.
The results are given in the format "average residuals (average residuals after sampling)" for each
scheme. For CPL-AdS the sampling period is the estimation period (epoch 0) as defined by the scheme.
The number of iterations per trial is 250 with 10000 independent trials per data point.}
\label{tab:func-c-results}
\end{table}

From Table \ref{tab:func-a-results}, \ref{tab:func-b-results} and \ref{tab:func-c-results} it is clear that
TS-SPL is the most efficient root solver among state-of-the-art schemes. This largely comes from the fact that it simultaneously learns
whether $g(x)$ is decreasing or increasing with $x$, as well as trying to locate the root $x^*$. In addition there
is the risk that the
sampling procedure that the other schemes apply to determine the direction $g(x)$ is increasing may conclude with the wrong answer. If this happens
then none of the schemes depending on the sampling will converge towards the root $x^*$.
Furthermore, an advantage of using TS-SPL is that it is can be applied to a wide range of functions
without regards to any local extrema residing in the function. This is unlike SA that shows
excellent performance only for monotonic functions as exemplified in Table \ref{tab:func-a-results}.

% ###################################################################
%
%  ######   #######  ##    ##  ######  ##       ##     ##  ######  ####  #######  ##    ## 
% ##    ## ##     ## ###   ## ##    ## ##       ##     ## ##    ##  ##  ##     ## ###   ## 
% ##       ##     ## ####  ## ##       ##       ##     ## ##        ##  ##     ## ####  ## 
% ##       ##     ## ## ## ## ##       ##       ##     ##  ######   ##  ##     ## ## ## ## 
% ##       ##     ## ##  #### ##       ##       ##     ##       ##  ##  ##     ## ##  #### 
% ##    ## ##     ## ##   ### ##    ## ##       ##     ## ##    ##  ##  ##     ## ##   ### 
%  ######   #######  ##    ##  ######  ########  #######   ######  ####  #######  ##    ## 
%
% ###################################################################
\section{Conclusions and Further Work}
\label{sec:conclusion}
In this paper, we investigated a novel reinforcement learning problem derived
 from the so-called "N-Door Puzzle". This puzzle has the fascinating property
 that it involves stochastic \emph{compulsive liars}. Feedback is erroneous on average,
 systematically misleading the decision maker. This renders traditional reinforcement learning (RL)
 based  approaches ineffective due to their dependency on
\emph{"on average"} correct feedback.

To solve the problem of deceptive feedback, we recast the problem as a particularly intriguing
 variant of the multi-armed bandit problem, referred to as the {\it
 Stochastic Point Location (SPL) Problem}. The decision maker is here only
 told whether the optimal point on a line lies to the ``left'' or to the
 ``right'' of a current guess, with the feedback being erroneous with
 probability $1-\pi$. Solving this problem opens up for optimization in
 continuous action spaces with both {\it informative} and {\it deceptive}
 feedback.
 
Our solution to the above problem, introduced in the present paper, is based on a novel compact and scalable Bayesian representation of the
 solution space. This model simultaneously captures both the location of the optimal
 point, as well as the probability of receiving correct feedback, $\pi$. We
 further introduced an accompanying Thompson Sampling guided Stochastic Point
 Location (TS-SPL) scheme for balancing exploration against exploitation. By
 learning $\pi$, TS-SPL supports  deceptive environments that are lying about the direction of the optimal point. 

The resulting scheme was applied to the Stochastic Point Location (SPL) problem and outperformed all of the Learning Automata driven methods. However, by enhancing the Soft Generalized Binary Search (SGBS) scheme with our Bayesian representation of the solution space, SGBS was able to outperform TS-SPL under informative feedback. For deceptive SPL problems, TS-SPL outperformed
all of the existing state-of-art schemes by several orders of magnitude, even when the latter schemes were supported by our Bayesian model.

We also applied TS-SPL to the Stochastic Root Finding Problem (SRFP). We demonstrated that SRFP can be seen as a deceptive problem, allowing TS-SPL to outperform existing dedicated state-of-art SRFP schemes by an order of magnitude.
Thus, TS-SPL can be considered state-of-the-art for both deceptive SPL and for the SRFP,
while yielding comparable results to the top performing schemes in the case of informative SPLs.

Despite the above performance gains, TS-SPL is based on Thompson Sampling, which is known to have a tendency
to over-explore high variance reward distributions \cite{lattimore2015optimally}. In future work, it is
therefore interesting to investigate mechanisms that eliminate or reduce this tendency, to further increase convergence speed.

\bibliographystyle{IEEEtran}
\bibliography{library}
\end{document}